%% file: main.tex
\documentclass[twoside]{article}

\usepackage[accepted]{aistats2020}

\usepackage[round]{natbib}

\usepackage[utf8]{inputenc} 
\usepackage[T1]{fontenc}    
\usepackage{url}            
\usepackage{booktabs}       
\usepackage{amsfonts}       
\usepackage{nicefrac}       
\usepackage{microtype}      

\usepackage{amsmath}
\usepackage{amsthm}
\usepackage{amssymb}
\usepackage{enumitem}

\usepackage[titletoc, toc]{appendix}

\usepackage{color}
\definecolor{mydarkblue}{rgb}{0,0.08,0.45}
\usepackage[colorlinks=true,allcolors=mydarkblue]{hyperref}

\usepackage{graphicx}
\graphicspath{ {./fig/} }

\newtheorem{theorem}{Theorem}

\newtheorem{proposition}[theorem]{Proposition}
\usepackage[bf]{caption}
\usepackage{float}
\usepackage{xcolor}

\include{preamble_maths}

\begin{document}

\runningtitle{Neural Decomposition: Functional ANOVA with Variational Autoencoders}

\twocolumn[

\aistatstitle{Neural Decomposition: \\ Functional ANOVA with Variational Autoencoders}

\aistatsauthor{Kaspar M\"artens \And  Christopher Yau} 

\aistatsaddress{University of Oxford \And Alan Turing Institute\\University of Birmingham\\University of Manchester} 

]

\begin{abstract}
Variational Autoencoders (VAEs) have become a popular approach for dimensionality reduction. However, despite their ability to identify latent low-dimensional structures embedded within high-dimensional data, these latent representations are typically hard to interpret on their own. Due to the black-box nature of VAEs, their utility for healthcare and genomics applications has been limited. 
In this paper, we focus on characterising the sources of variation in Conditional VAEs. Our goal is to provide a \emph{feature-level} variance decomposition, i.e. to decompose variation in the data by separating out the marginal additive effects of latent variables $\boldz$ and fixed inputs $\boldc$ from their non-linear interactions. We propose to achieve this through what we call \emph{Neural Decomposition} -- an adaptation of the well-known concept of functional ANOVA variance decomposition from classical statistics to deep learning models.
We show how identifiability can be achieved by training models subject to constraints on the marginal properties of the decoder networks. We demonstrate the utility of our Neural Decomposition on a series of synthetic examples as well as high-dimensional genomics data.
\end{abstract}

\setlength{\abovedisplayskip}{2pt}
\setlength{\belowdisplayskip}{2pt}

\section{Introduction}

\begin{figure*}[!t]
    \centering
    \includegraphics[width=\textwidth]{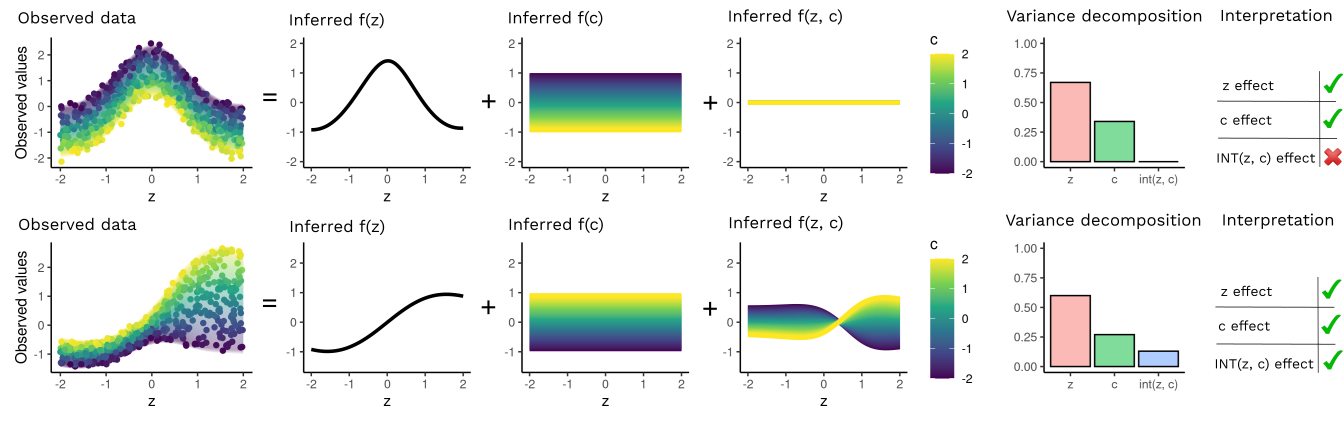}
    \vspace{-6mm}
    \caption{Neural Decomposition illustrated. Given high-dimensional data (shown for two selected features) and covariate information $\boldc \in \mathbb{R}$ (in colour coding), we would like to learn a low-dimensional representation $\boldz \in \mathbb{R}$ (on $x$-axis) together with a decomposition into marginal $\boldz$ and $\boldc$ effects and their interaction $f(\boldz, \boldc)$. ``Variance decomposition'' would quantify the variance attributed to each component and aid interpretability.}
    \vspace{-2mm}
    \label{fig:neural_decomp}
\end{figure*}

Dimensionality reduction is often required for the analysis of complex high-dimensional data sets in order to identify low-dimensional sub-structures that may give insight into patterns of interest embedded within the data. 
Recently there has been particular interest in applications of Variational Autoencoders (VAEs) \citep{kingma_auto-encoding_2014} for generative modelling and dimensionality reduction. VAEs are a class of probabilistic neural-network-based latent variable models. Their neural network (NN) underpinnings offer considerable modelling flexibility while modern programming frameworks (e.g.\ TensorFlow and PyTorch) and computational techniques (e.g.\ stochastic variational inference) provide immense scalability to large data sets, including those in genomics \citep{lopez_deep_2018, eraslan_single-cell_2019, martens_basisvae_2020}.
The Conditional VAE (CVAE) \citep{sohn_learning_2015} extends this model to allow incorporating side information as additional fixed inputs. 

The ability of (C)VAEs to extract latent, low-dimensional representations from high-dimensional inputs is a strength for predictive tasks. For example (C)VAEs have been particularly popular as generative models for images. However, when the objective is to understand the contributions of different inputs to components of variation in the data, the black box decoder is insufficient to permit this. This is particularly important for \emph{tabular} data problems where individual features might correspond to actual physical quantities (e.g. expression of a gene or a physiological measurement) and we are interested in how the variability of the multivariate output response is driven by specific changes in the inputs (where inputs may be either latent variables or observed covariates).

Functional analysis of variance (functional ANOVA, or fANOVA) are a class of statistical models that uniquely decompose a functional response according to the main effects and interactions of various factors \citep{sobol_sensitivity_1993, ramsay_functional_1997}. For example, smoothing spline ANOVA models \citep{gu_smoothing_2013} express the effects as linear combinations of some underlying basis functions, and the coefficients on these basis functions are  chosen to minimise a loss function balancing goodness of fit with a measure of smoothness of the fitted functions. 
More recent treatments have proposed functional ANOVA decompositions via tree-based models \citep{lengerich_purifying_2019}, and within a Bayesian nonparametric framework using Gaussian Processes \citep{kaufman_bayesian_2010}.

In this paper, we propose the notion of \emph{neural decomposition} -- the integration of functional ANOVA and deep neural networks for dimensionality reduction and variance decomposition. Specifically, we develop and explore a class of (C)VAE models where the \emph{decoding} network is specifically implemented to explain feature-level variability in terms of additive and interaction effects between the latent variables and any additional covariate information (Figure \ref{fig:neural_decomp}). 

Thus, neural decomposition can be seen from two perspectives:
\begin{itemize}
    \item From the classical statistics perspective, it is a scalable \emph{neural} adaptation of the functional ANOVA decomposition, where additional latent variables $\boldz$ have been introduced to capture low-dimensional structure within high-dimensional data and nonlinearities implemented via deep neural nets.
    \item From the deep learning perspective, it is an extension of the (C)VAE framework, where the decoder has a decomposable additive structure to aid interpretability on the level of individual features.
\end{itemize}
Crucially, in order for this class of models to be useful, identifiability must be enforced to ensure the expressive power of each neural network component does not absorb variation that should be explained by others. Inference for our model relies on the imposition of functional constraints which we implement as \emph{constrained} optimisation in the weight-space of neural networks.

\section{Background}

\textbf{VAEs:} Variational Autoencoders are constructed based on particular a latent variable model structure. Let $\boldy \in {\cal Y}$ denote a data point in some potentially high-dimensional space and $\boldz \in {\cal Z}$ be a vector of associated latent variables typically in a much lower dimensional space. We assume that the prior $p(\boldz)$ is from a family of distributions that is easy to sample from - we will assume $\boldz \sim \mathcal{N}(\bf{0}, \bf{I})$ throughout. Now suppose $f^\theta(\boldz)$ is a family of deterministic functions, indexed by parameters $\theta \in \Theta$, such that $f : {\cal Z} \times \Theta \rightarrow {\cal Y}$. 
Given data points $\boldy_{1:N} = \{ \boldy_1, \dots, \boldy_N \}$, the marginal likelihood is given by 
$
    p(\boldy_{1:N}) = \prod_{i=1}^N  \int p_\theta(\boldy_i|\boldz_i) p(\boldz_i) d\boldz_i .
$
We will assume a Gaussian likelihood, i.e.\ $\boldy_i | \bold z_i, \theta \sim \mathcal{N}(f^\theta(\boldz_i), \sigma^2)$.
In the VAE, these deterministic functions $f^{\theta}$ are given by deep neural networks (NNs). 
Posterior inference in a VAE is carried out via \emph{amortised} variational inference, i.e.\ with a parametric inference model $q_\phi(\boldz_i|\boldy_i)$ where $\phi$ are variational parameters. 
Application of standard variational inference methodology, see e.g.\ \citep{blei_variational_2017}, leads to a lower bound on the log marginal likelihood, i.e.\ the ELBO of the form:
\setlength{\abovedisplayskip}{3pt}
\setlength{\belowdisplayskip}{3pt}
\begin{align*}
    \mathcal{L}^{\theta, \phi} = \sum_{i=1}^N \mathbb{E}_{q_{\phi}(\boldz_i | \boldy_i)} [\log p_\theta(\boldy_i | \boldz_i)] 
    - \text{KL}(q_{\phi}(\boldz_i|\boldy_i) || p(\boldz_i)) 
\end{align*}

In the VAE, the variational approximation $q_\phi(\boldz_i|\boldy_i)$ is referred to as the \emph{encoder} since it encodes the data $\boldy$ into the latent variable $\boldz$ whilst the \emph{decoder} refers to the generative model $p_\theta(\boldy|\boldz)$ which decodes the latent variables into observations. 

Training a VAE seeks to optimise both the model parameters $\theta$ and the variational parameters $\phi$ jointly using stochastic gradient ascent. This typically requires a stochastic approximation to the gradients of the variational objective which is itself intractable. Typically the approximating family is assumed to be Gaussian, i.e. $q_\phi(\boldz_i|\boldy_i) = \mathcal{N}(\boldz_i|  \mu_{\phi}(\boldy_i), \sigma^2_{\phi}(\boldy_i))$, so that a reparametrisation trick can be used \citep{kingma_auto-encoding_2014, rezende_stochastic_2014}.

\setlength{\abovedisplayskip}{2pt}
\setlength{\belowdisplayskip}{2pt}

\textbf{Conditional VAEs:} The Conditional VAE \citep{sohn_learning_2015} augments the VAE by conditioning the generative model on additional inputs $\boldc$, i.e.\ the VAE generative model $\boldy_i = f^\theta(\boldz_i) + \epsilon_i$ is now replaced by $\boldy_i = f^\theta(\boldz_i, \boldc_i) + \varepsilon_i$
which requires a minor alteration to the ELBO
\begin{align*}
    \mathcal{L}^{\theta, \phi} = \sum_{i=1}^N \mathbb{E}_{q_{\phi}(\boldz_i | \boldy_i, \boldc_i)} [\log p_\theta(\boldy_i | \boldz_i, \boldc_i)] \\
    - \text{KL}(q_{\phi}(\boldz_i|\boldy_i, \boldc_i) || p(\boldz_i)) .
\end{align*}
Conditioning on the extra inputs can greatly increase the expressiveness of the VAE when applied to multimodal data where the indicator driving the multimodality is measurable. 
For the genomics applications we consider, we use this mechanism to incorporate \emph{known covariates} as part of the model. 



\section{Neural Decomposition}

Our goal is to equip (C)VAEs with feature-level interpretability that would let us characterise the sources of variation for individual features. We aim to achieve this by directly embed \emph{decomposable} structure within this latent variable model. 
Specifically, we propose to perform functional ANOVA decomposition as part of the \emph{decoder} network within the (C)VAE, in order to simultaneously perform dimensionality reduction as well as obtain a feature-level variance decomposition. 

For illustration purposes, we consider the special case $\boldz, \boldc \in \mathbb{R}$. Here we want to decompose the decoder network $f^\theta(\boldz_i, \boldc_i)$ to extract additive marginal and interaction effects as follows
\begin{align}
    f^\theta(\boldz_i, \boldc_i) = f_0^\theta + f_z^\theta(\boldz_i) + f_c^\theta(\boldc_i) + f_{zc}^\theta(\boldz_i, \boldc_i)
    \label{eq:cvae_decoder_anova}
\end{align}
as illustrated in Figure~\ref{fig:neural_decomp}. 
However, note that for this functional decomposition to have a meaningful interpretation, identifiability must be enforced.

Traditionally, in functional ANOVA the functions $f_z^\theta, f_c^\theta, f_{zc}^\theta$ would be represented by either linear mappings, smoothing splines \citep{gu_smoothing_2013} or Gaussian Processes \citep{kaufman_bayesian_2010}. Here, we propose to use deep neural networks instead -- a \emph{Neural Decomposition} (ND). This would enable separating main effects from complex interactions while otherwise maintaining the flexibility of deep generative models, and fast, approximate inference via variational methods. This decoding structure has been illustrated in Fig~\ref{fig:ND_schema}(C). 

Next, we formulate the conditions for the decomposition to be unique, and discuss how such constraints can be fulfilled in practice. 

\subsection{Identifiable Neural Decomposition}

\begin{figure*}[!t]
    \centering
    \includegraphics[width=0.9\textwidth]{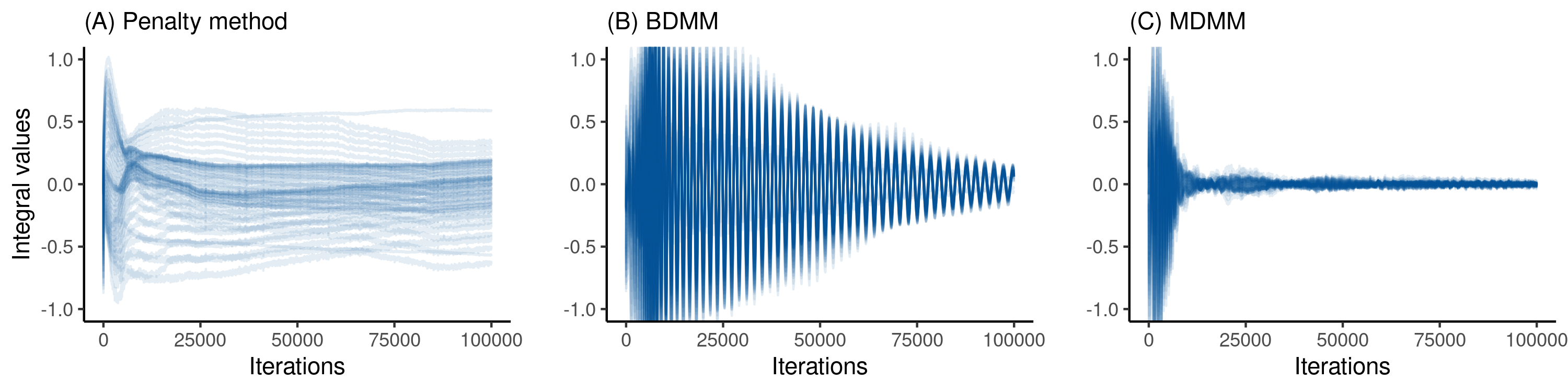}
    \vspace{-3mm}
    \caption{Different strategies for optimisation under constraints, showing the traces for $\int f^{\theta}(x_1, x_2) dx_2$ on a grid of $x_1$ values (each line corresponds to one grid point) over 100~000 iterations. (A) The constraints have not been fulfilled by the penalty method with a fixed $c$. (B) BDMM exhibits oscillating behaviour and integrals are slowly converging towards zero, whereas the hybrid (C) MDMM which combines the two penalties A and B (using the same $c$ and same learning rate $\eta$) leads to optimisation which results in $\approx$ 0 integral values much more quickly. }
    \label{fig:optimisation}
    \vspace{-4mm}
\end{figure*}

For a more general formulation, let us denote the latent and fixed inputs collectively by $\boldx := (\boldz, \boldc)$ with dimensionality $D := \text{dim}(\boldx_i)$\footnote{Whilst we focus on the CVAE setting in this paper, the decomposition is more generally applicable. Thus in this section, we treat all inputs as fixed.}. We would like to decompose as follows
\begin{equation} \label{eq:naive_decomposition}
    f_0^{\theta} + \sum_k f_k^{\theta}(x_{ik}) + \sum_{k, l} f^{\theta}_{kl}(x_k, x_l) + \ldots + f^{\theta}_{1, ..., D}(\mathbf{x})
\end{equation}
where all functions $f_k^{\theta}, f_{k,l}^{\theta}, \ldots, f^{\theta}_{1, ..., D}$ are parameterised by neural networks.
We note that without any additional constraints on the neural networks the above decomposition \eqref{eq:naive_decomposition} is unidentifiable. This is because the functional subspaces $f_{\mathcal{I}}$ corresponding to different index sets ${\mathcal{I}}$ can all be seen as functions defined on the same input space $\mathbb{R}^D$, being constant in the rest of coordinates, and these subspaces are overlapping:

\begin{proposition}
Let $f_i^{\theta}, f_{ij}^{\theta}, \ldots, f_{1, ..., D}^{\theta}$ be neural networks with the same architecture, i.e.\ assuming that the networks only differ in the number of inputs\footnote{
    Assuming a fully connected first layer with $H$ hidden units, i.e.\ that the first transformation applied to inputs $\boldx_\calI$ is $\boldW \boldx_\calI + \boldb$ where $\boldW \in \R^{H \times \text{dim}(\boldx_\calI)}, \boldb \in \R^H$, by ``the same architecture'' we mean that $\text{dim}(\boldx_\calI)$ is the only element that is allowed to vary across $f^\theta_\calI$.
}. Then for any two disjoint sets of indices $\mathcal{I}$ and $\mathcal{J}$ the functional subspace $\{f_{\mathcal{I}}^{\theta}(\boldx_{\mathcal{I}}): \theta \in \mathbb{R}^{|\theta|}\}$ is strictly a subset of the functional subspace $\{f_{\mathcal{I}, \mathcal{J}}^{\theta}(\boldx_{\mathcal{I}}, \boldx_{\mathcal{J}}): \theta \in \mathbb{R}^{|\theta|}\}$.
\end{proposition}
\textit{Proof: weights in the first layer corresponding to inputs $\boldx_{\mathcal{J}}$ can be set to zero, eliminating the effect of $\boldx_{\mathcal{J}}$.
}

As a result, without any additional constraints decomposition~\eqref{eq:naive_decomposition} is not meaningful: it can be used for predictive purposes, but the relative contribution of different terms has no direct interpretation since higher-order interactions can \emph{absorb} variability that could be explained by main effects or low-order interactions. 

To turn this into an \emph{identifiable} learning problem, we need to introduce functional constraints. As in functional ANOVA, we introduce the integral constraints to constrain the marginal effects of every neural network $f^{\theta}_{\mathcal{I}}(\boldx)$ to be zero. This can be formalised as follows:
\begin{proposition}
Let neural networks $f_i^{\theta}, f_{ij}^{\theta}, \ldots, f_{1, ..., D}^{\theta}$ be such that they satisfy the following integral constraints
\begin{equation*}
    \int f_{\calI} (\boldx_\calI) d x_i = 0 \text{ for all } i \in \calI
\end{equation*}
for all neural networks $f_{\calI}$ in~\eqref{eq:naive_decomposition}, i.e.\ for every index set $\calI \subset  \{1, \ldots, D\}$.
Then for any $\mathcal{I} \cap \mathcal{J} = \varnothing$ the functional subspaces corresponding to $f_{\mathcal{I}}^{\theta}$ and $f_{\mathcal{I}, \mathcal{J}}^{\theta}$ do not overlap any more (apart from the constant zero function). Furthermore, these functional subspaces are orthogonal in $L_2$. 
\end{proposition}
Both of these properties (no overlap and orthogonality) are a direct consequence of the integral constraints. The general proof follows derivations as in previous works, see e.g. \citep{sobol_sensitivity_1993}.
We note that the former (no overlap) is sufficient for identifiability, but the latter (orthogonality) leads to an easily interpretable variance decomposition
\begin{align*}
    \Var(f_0^{\theta}) &+ \sum_k \Var(f_k^{\theta}(x_{ik})) + \\
    &+ \sum_{k, l} \Var(f^{\theta}_{kl}(x_i, x_j))
    + \ldots + \Var(f^{\theta}_{1, ..., D}(\mathbf{x}))
\end{align*}
For the special case of a two-dimensional input $(x_1, x_2)$, i.e.\ when the decomposition consists of $f_0^{\theta}, f_1^{\theta}, f_2^{\theta}, f_{12}^{\theta}$, the above leads to the following integral constraints: 
\begin{itemize}[nolistsep]
    \item $\int f_1^{\theta}(x_1) dx_1 = 0$ and $\int f_2^{\theta}(x_2) dx_2 = 0$
    \item $\int f_{12}^{\theta}(x_1, x_2) dx_1 = 0$ for all $x_2$ 
    \item $\int f_{12}^{\theta}(x_1, x_2) dx_2= 0$ for all $x_1$.
\end{itemize}

\subsection{Inference under integral constraints}

We now return to our original goal which is understanding the sources of variation in (C)VAEs. As in eq~\eqref{eq:cvae_decoder_anova}, our goal is to learn a decomposition as part of the \emph{generative} model -- in the VAE framework it corresponds to decomposing the \emph{decoder}\footnote{We note that we aim to make the \emph{functional decomposition} (as opposed to the entire (C)VAE) identifiable.}. 
To obtain an identifiable neural decomposition, we need to enforce the above integral constraints for the decoding networks. 

This is a constrained optimisation problem, which is non-trivial to solve in the context of deep learning where state-of-the-art off-the-shelf optimisation techniques, such as Adam \citep{kingma_adam:_2014}, are typically implemented for unconstrained problems only. 
Thus alternative strategies have to be considered. 
We turn to the \emph{Augmented Lagrangian} method, also known as the method of multipliers \citep{hestenes_multiplier_1969, powell_method_1969}, to enforce such constraints. We refer to \citep{platt_constrained_1988} for an adaptation to neural networks. 


\textbf{Augmented Lagrangian for a single constraint:}
For illustration, we first consider the case where our decoder $f^{\theta}(x)$ has a univariate input $x$, and we want to optimise the ELBO subject to a constraint $\int f^{\theta}(x) dx = 0$, i.e.\ we want to restrict the $f^{\theta}$ to a subspace such that $\int f^{\theta}(x) dx = 0$. 
To enforce this constraint, we will augment the ELBO with additional penalty term(s) which will be equal to zero when the integral constraints are fulfilled. The resulting objective function is not necessarily a lower bound, but reduces to the ELBO once the constraints become fulfilled during optimisation.

One such approach for would be the \emph{penalty method}, i.e.\ to incorporate 
a penalty term $c \cdot \left( \int f^{\theta}(x) dx \right)^2$ with a fixed penalty $c$. This approach has the disadvantage that for a fixed value of $c$ we do not have any guarantees that constraints would be fulfilled exactly.

Alternatively one could introduce a penalty $\lambda \int f^{\theta}(x) dx$ where now $\lambda$ is treated as a parameter. This is analogous to the use of Lagrange multipliers, and following the terminology of \citep{platt_constrained_1988}, we refer to this as the Basic Differential Multiplier Method (BDMM). 
Instead of gradient updates $\lambda^{t+1} = \lambda^t - \eta 
\left( \int f^{\theta}(x) dx \right)$, BDMM would follow the opposite direction $\lambda^{t+1} = \lambda^t + \eta \left( \int f^{\theta}(x) dx \right)$ when optimising $\lambda$. Platt and Barr showed that this corresponds to optimisation behaviour where the system undergoes damped oscillation. 

We have empirically compared the behaviour of the penalty method and the BDMM approach on a synthetic problem with two-dimensional inputs, as shown in Figure~\ref{fig:optimisation}A. Using a fixed penalty does not necessarily lead to fulfilled constraints, whereas the oscillating behaviour of the BDMM leads to the integrals converging towards zero.

Finally, these two penalty terms can be combined, resulting in a hybrid constrained optimisation objective
$$
    \min_{\theta, \phi} \left\{ 
        - \mathcal{L}^{\theta, \phi} + \lambda \int f^{\theta}(x) dx + c \left( \int f^{\theta}(x) dx \right)^2
    \right\}
$$
where $\lambda$ is optimised, $c$ is a fixed constant, and $\mathcal{L}^{\theta, \phi}$ is the ELBO for the (C)VAE. 
This scheme, the Modified Differential Multiplier Method (MDMM), results in more robust behaviour, both from a theoretical perspective \citep{platt_constrained_1988} as well as supported by empirical evidence as shown in our Figure~\ref{fig:optimisation}. Furthermore, we have observed that replacing a fixed $c$ with a sequence $c^{1} \le  \cdots \le c^{T}$ can empirically lead to even faster convergence.

\textbf{Enforcing multiple identifiability constraints:} Next we discuss how to handle multiple constraints, illustrating this on a special case with a two-dimensional input $(x_1, x_2)$. 
In order to satisfy the constraints $\int f_{12}^{\theta}(x_1, x_2) dx_2$ for every value $x_1$ in some interval, we need to introduce a Lagrange multiplier $\lambda_{x_1}(x_1)$ which is now indexed by a continuous-valued $x_1$. The additional penalty corresponding to this term will be $\int \lambda_{1}(x_1) \left( \int f_{12}^{\theta}(x_1, x_2) dx_1 \right) dx_2$. Similarly, the ELBO will also be augmented by $\int \lambda_{2}(x_2) \left( \int f_{12}^{\theta}(x_1, x_2) dx_2 \right) dx_1$ in addition to penalty terms $\lambda_{1} \int f_{1}^{\theta}(x_1) dx_1$ and $\lambda_{2} \int f_{2}^{\theta}(x_2) dx_2$. 
In practice, we can choose to estimate these integrals using either quadrature or Monte Carlo estimates.

A natural question that arises is how do we know whether the constraints have been (approximately) satisfied. We propose to approach this as follows: we establish a desired tolerance threshold $\varepsilon$ and evaluate the integrals after optimisation to make sure that all NNs have been constrained to the desired functional subspaces within the desired tolerance. 
Note that all the additional penalty terms need to be evaluated during training time only. This is important, because once the model has been trained, 
it can be used for prediction without any additional costs at test time.

\subsection{Sparse Neural Decomposition}

For interpreting what the decomposed (C)VAE has learnt, in addition to obtaining the variance decomposition it is also of practical interest to \emph{detect} the presence or absence of dependence on certain input coordinates, e.g.\ detect groups of genes which depend purely on $\boldz$ and not on $\boldc$. This would make it easier to interpret. 

The feature-level variance decomposition from ND does not explicitly identify such groups by default. One approach would be to apply an \textit{ad hoc} thresholding to decide which effects are zero. 
For a more principled probabilistic approach, in this section we introduce Bernoulli random variables for every decoder network and for every data dimension. For the special case when $\boldz, \boldc \in \mathbb{R}$, we would introduce $s_z^{(j)}, s_c^{(j)}, s_{zc}^{(j)} \sim \text{Bernoulli}(p_0)$ for every feature $j=1, \ldots, P$, each indicating the presence of $f_z^{\theta}, f_c^{\theta}$ and $f_{zc}^{\theta}$ non-zero effects for feature $j$. I.e.\ we define the decoder to have the following structure
\begin{align*}
    \boldy_i | \boldz_i, \boldc_i, \theta, \bolds = \bolds_c f_c^{\theta}(\boldc_i) + \bolds_z f_z^{\theta}(\boldz_i) + \bolds_{cz} f_{cz}^{\theta}(\boldz_i, \boldc_i) + \varepsilon_i
\end{align*}
where $\bolds_c, \bolds_z, \bolds_{cz}$ have the same dimensionality as observations $\boldy$.
To implement variational inference via the reparameterisation trick, we use the continuous relaxation of Bernoulli random variables \citep{maddison_concrete_2017, jang_categorical_2017}. We specify a prior $p(\bolds) = \text{RelaxedBernoulli}(p_0)$ for all $\bolds_c, \bolds_z, \bolds_{cz}$ and approximating distributions $q(\bolds)=\text{RelaxedBernoulli}(\boldu)$ where $\boldu$ are variational parameters. We use $p_0=0.1$ throughout.

\begin{figure}[!th]
    \centering
    \includegraphics[width=\columnwidth]{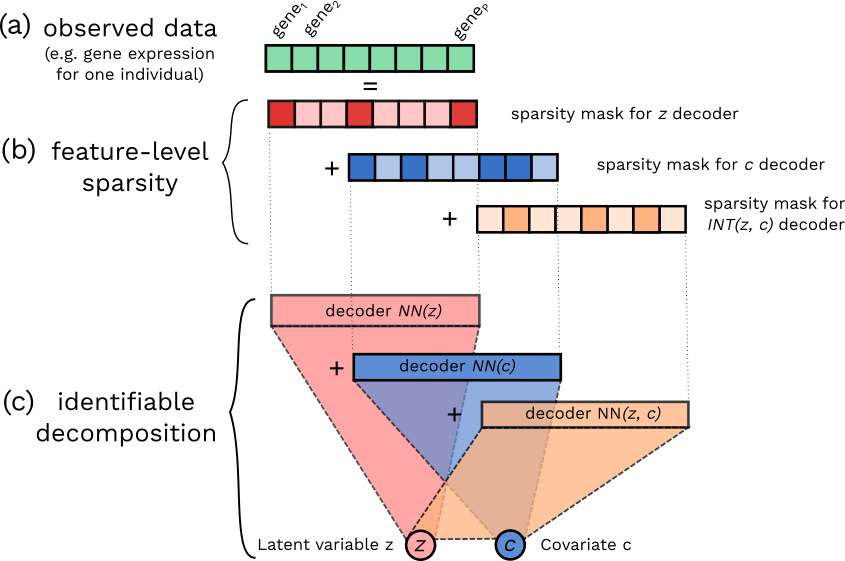}
    \caption{Diagram of the Neural Decomposition decoding model. For every gene in panel (a), we learn an identifiable decomposition of decoder networks $f_z^\theta(\boldz_i), f_c^\theta(\boldc_i)$ and $f_{zc}^\theta(\boldz_i, \boldc_i)$ as shown in panel (c), which are then elementwise multiplied with the respective Bernoulli sparsity masks in panel (b) for easier interpretability.}
    \vspace{-4mm}
    \label{fig:ND_schema}
\end{figure}

The integration of the sparsity priors and the neural decomposition model is summarised in Figure \ref{fig:ND_schema}.

\section{Related work}

There has been substantial interest towards \emph{explaining} black-box models. A common strategy is to approximate the neural network with a simpler, interpretable model. Such approaches can be divided into \emph{global} \citep{tan_learning_2018} and \emph{local} explanations \citep{ribeiro_why_2016, lundberg_unified_2017}. 
As opposed to such \emph{post hoc} explanation methods, Neural Decomposition has \emph{decomposable} structure built in as part of the model. 

From the application perspective, \citet{martens_decomposing_2019} have considered a similar problem setting, but they relied on Gaussian Processes (GPs) as opposed to neural networks. While in the GP framework we can enforce functional constraints elegantly in closed form, its scalability properties are different. We have empirically found ND to be much more scalable to high-dimensional data, both in terms of compute and memory (see Supplementary for details). 

\begin{figure*}[!t]
    \centering
    \includegraphics[width=\textwidth]{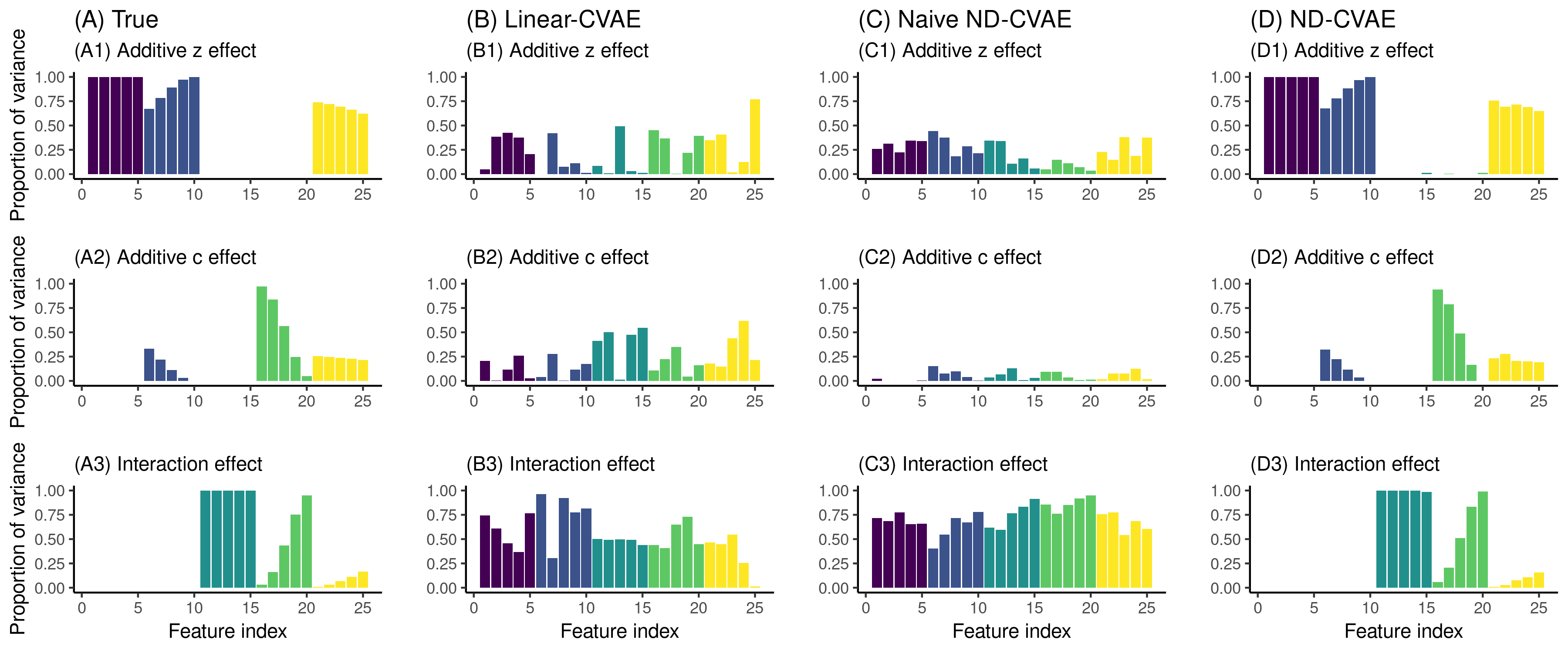}
    \vspace{-5mm}
    \caption{The true variance decomposition $\Var(f_{z})$, $\Var(f_{c})$, $\Var(f_{zc})$ on a synthetic data set shown in (A1-A3) for all 25 features (which have been grouped in blocks of 5 for visual aid), compared to the inferred decompositions by (B) linear-CVAE, (C) unidentifiable ND, and (D) identifiable ND.}
    \label{fig:toy_exprs}
    \vspace{-2mm}
\end{figure*}

\section{Experiments}

The value of ND lies in \emph{scalable} and \emph{identifiable} functional decompositions. In real data, however, the true underlying functional decompositions are unknown, thus we are unable to quantify the correctness of the inferred decompositions on real data. For this reason, we seek to carefully characterise the behaviour of ND in a controlled setting before considering the real-life genomics example. 
Our PyTorch implementation of ND is available in \url{https://github.com/kasparmartens/NeuralDecomposition}.

\subsection{Synthetic data}

We first investigate the behaviour of ND in a controlled setting. We generated synthetic data with a goal to mimic patterns in real gene expression data, thus including linear relationships, monotone warpings as well as non-monotone non-linear dependency structures (see Supplementary for details). We applied three variants of ND: 
\begin{enumerate}[nolistsep, label=(\roman*)]
    \item ND-CVAE with a linear decoder
    \item ND-CVAE with a non-linear decoder without identifiability constraints
    \item our full implementation
\end{enumerate}


Figure~\ref{fig:toy_exprs} illustrates the inferred variance decompositions for 25 features with varying dependency structures (e.g. the first five features exhibit a pure additive $\boldz$ effect). Neither the linear-CVAE (panel (B)) nor the ND-CVAE without constraints (panel (C)) have been able to identify the correct decomposition: the former is unable to capture non-linearities in the data due to its restrictive modelling assumptions, and the latter suffers from unidentifiability which makes its inferred variance decomposition arbitrary. 

To demonstrate that our approach is not restricted to a univariate $\boldz$, we have included an experiment using a two-dimensional latent space (see batch correction example in Supplementary).

\subsection{CelebA}

The motivation behind ND is understanding \emph{tabular} (non-image) data -- this is the reason why we learn the decomposition on the level of observed features. However, the ND methodology is generally applicable to a variety of large-scale problems, and in this section we demonstrate ND on CelebA data \citep{liu_deep_2015}\footnote{Our interest is \emph{not} to tune ND to develop a state-of-the-art model for images (for this purpose, we would e.g.\ use CNNs within our CVAE), but instead to demonstrate ND on an easy-to-visualise high-dimensional use case.}.

\subsubsection{Pixel-level decomposition}

\begin{figure}[!h]
    \centering
    \includegraphics[width=0.95\columnwidth]{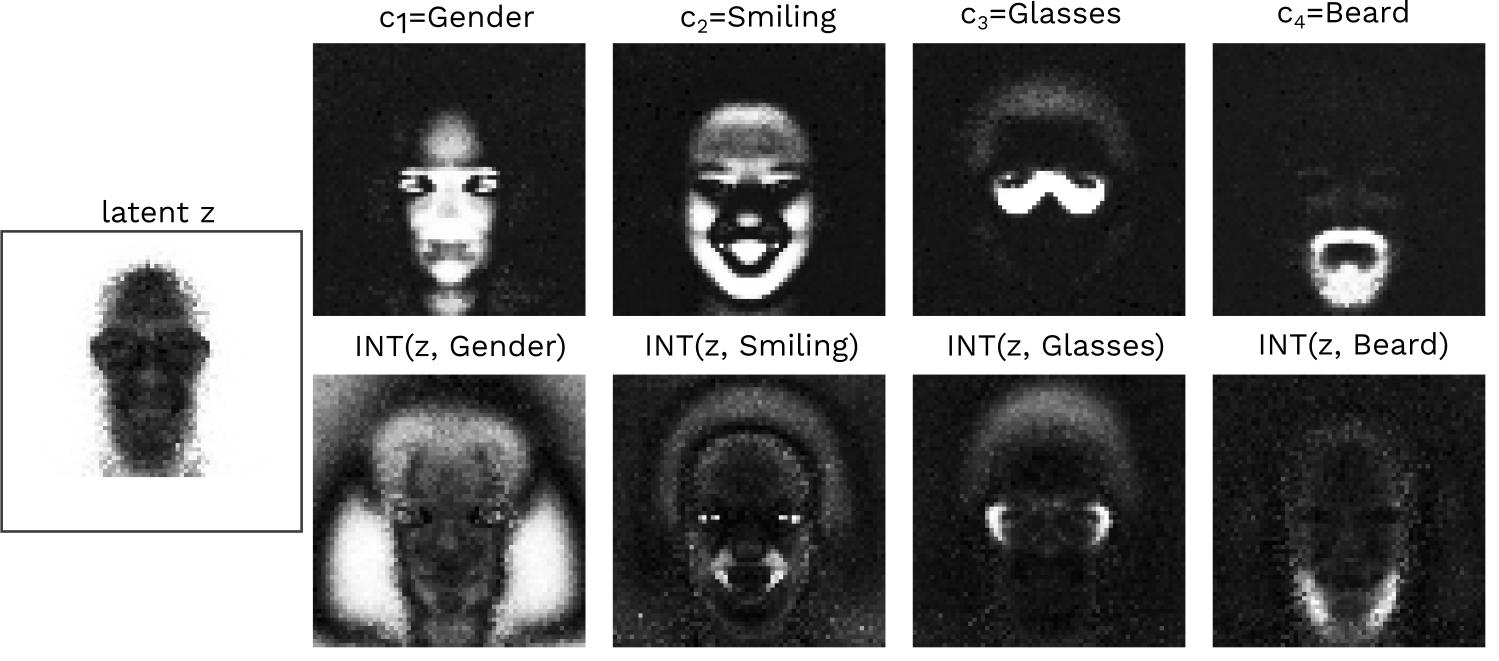}
    \vspace{-2mm}
    \caption{Inferred sparsity masks on CelebA data.}
    \vspace{-2mm}
    \label{fig:celeba_sparsity}
\end{figure}

To be able to capture subtle effects like ``glasses'' and ``beard'' which are typically not captured in the inferred latent space of the VAE (see e.g.\ empirical results in \citep{kim_disentangling_2018}), we include these as fixed covariates. Thus, we consider a CVAE with four covariates $c_1$=Gender, $c_2$=Smiling, $c_3$=Glasses, $c_4$=Beard and a 1D latent variable $z$. 
By performing a pixel-level decomposition with our ND with a functional form $f_z^{\theta}(z) + \sum_{k} f_{c_k}^{\theta}(c_k) + \sum_{k} f_{zc_k}^{\theta}(z, c_k)$ where $k \in \{1, 2, 3, 4\}$, thus capturing the marginal additive effects of $z$ and $\{c_k\}$ as well as their interactions as part of the decoding structure. The inferred pixel-level sparsity masks in Figure~\ref{fig:celeba_sparsity} help us interpret what each of the decoder networks has learnt. Note that $z$ captures how dark/light the background is, whereas the additive effects of covariates affect only relevant parts of the faces (e.g.\ glasses or beard). The interaction effects between $z$ and $c_k$ are relatively small: they mostly affect pixels that interact with the overall background darkness, e.g.\ long hair for women, or the outer borders of glasses and beard.

\subsubsection{Improved extrapolation} 

\begin{figure}[!h]
    \centering
    \includegraphics[width=0.95\columnwidth]{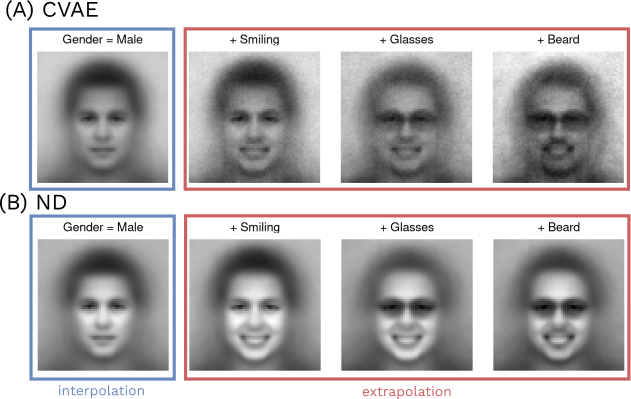}
    \vspace{-2mm}
    \caption{Generalization to unseen subgroups.}
    \vspace{-2mm}
    \label{fig:celeba_extrapolation}
\end{figure}

Despite the empirical success of deep generative models, their generalization (e.g. to under-represented subpopulations) in the presence of sampling bias is still an active area of research \citep{zhao_bias_2018}. Here, we demonstrate how additive structure in the decoder can improve the extrapolation properties of the CVAE. We use CelebA data to mimic sampling bias as follows: we exclude all images of all smiling men from the training set, and at test time, we visualise predictions, both for CVAE and ND-CVAE, as shown in Figure~\ref{fig:celeba_extrapolation}, 
\begin{itemize}[nolistsep]
    \item First, using $z=0$, Gender=Male, Smiling=False, Glasses=False, Beard=False, i.e.\ a scenario that was observed during training
    \item New scenarios Smiling=True (2nd panel), with additionally Glasses (3rd) and Beard (4th panel)
\end{itemize}

It is not surprising that CVAE has not been able to generate clear images for unseen subgroups: the decoder has not been trained on these particular inputs. However, the additive structure in ND has led to visually better quality images. 
The same is indicated by the loglikelihoods on held-out images of smiling men: for CVAE (-3181 $\pm$ 630) these are lower than for ND (-2497 $\pm$ 319).

\subsection{Gene expression}

\begin{figure}[!th]
    \centering
    \includegraphics[width=0.475\textwidth]{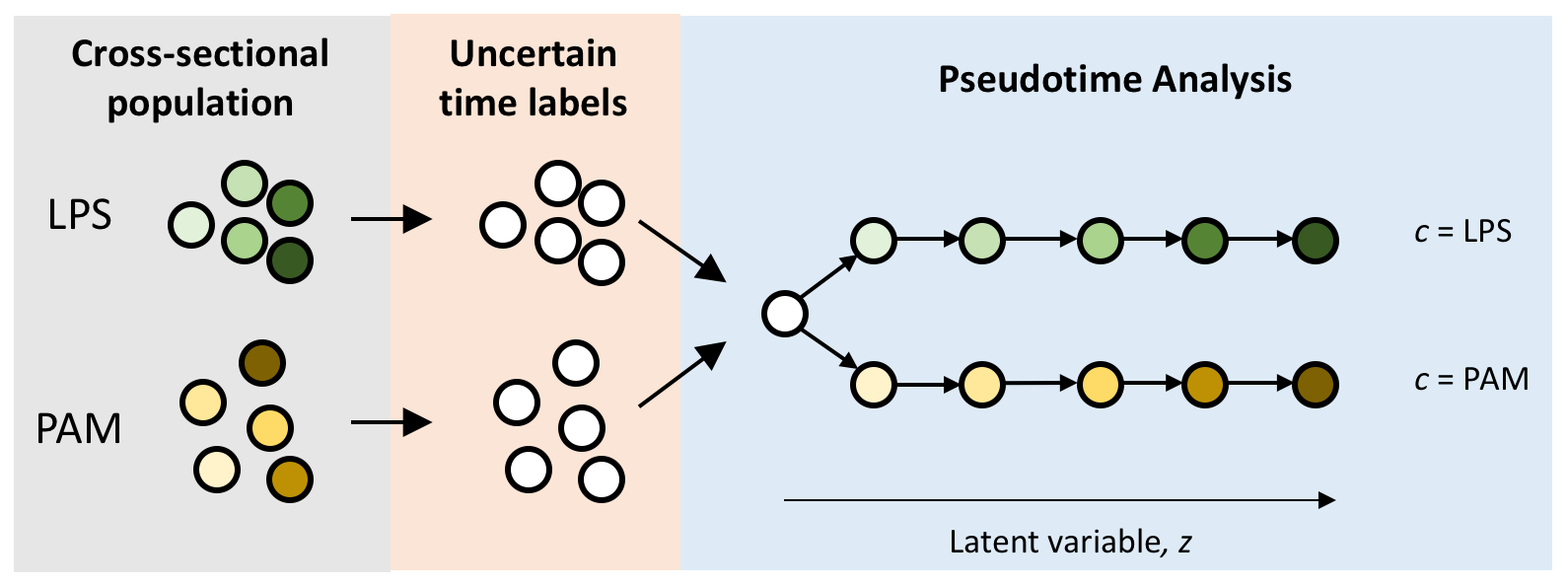}
    \includegraphics[width=0.4\textwidth]{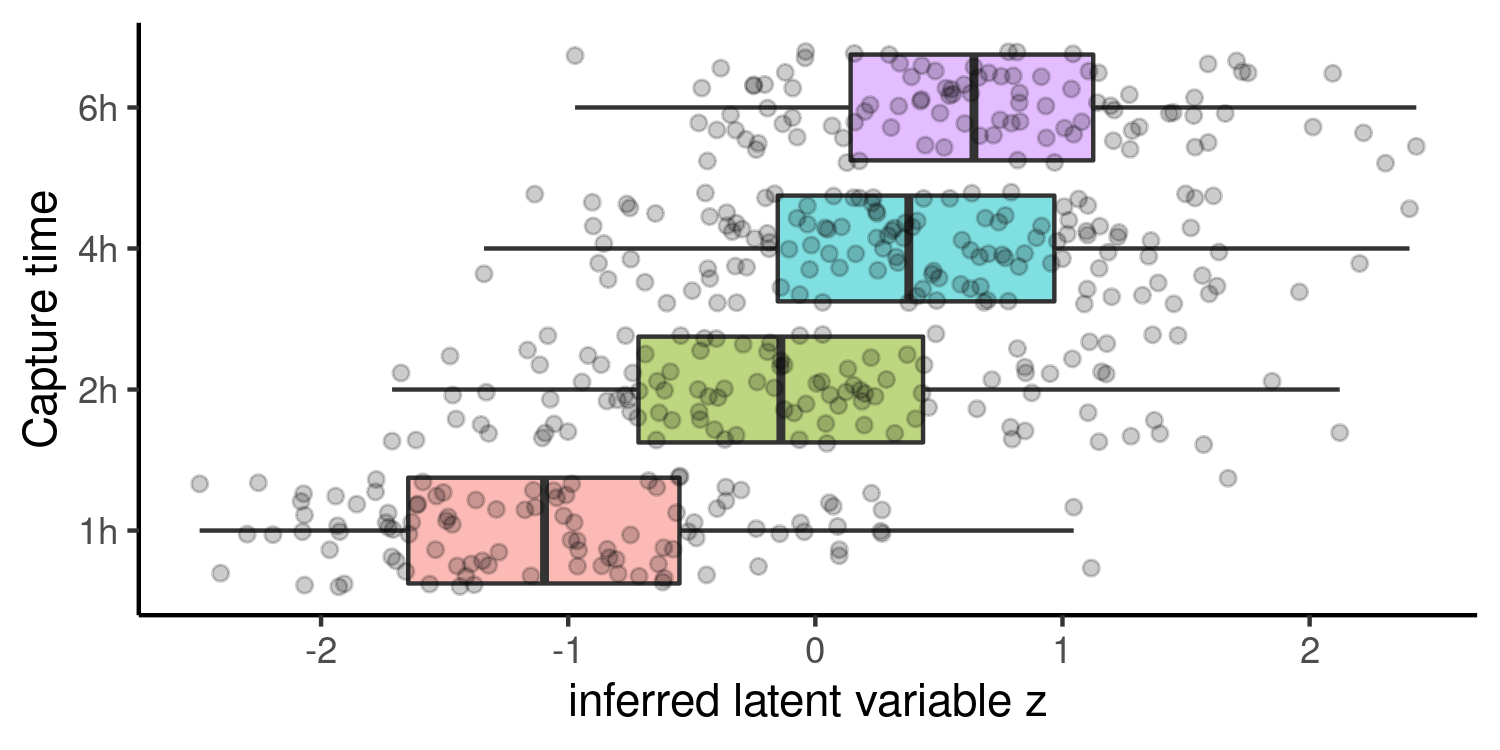}
    \vspace{-3mm}
    \caption{Above: Schema illustrating pseudotime analysis approach to mouse dendritic cells. Below: Inferred $z$ (pseudotime) correlates with capture time.}
    \vspace{-2mm}
    \label{fig:lpspam}
\end{figure}

We next examine a publicly available time-series single-cell RNA-seq (scRNA-seq) data set of bone marrow derived dendritic cells responding to particular stimuli \citep{shalek_single-cell_2014}. In this experiment, cells were exposed to either LPS (a component of Gram-negative bacteria) or PAM (a synthetic mimic of bacterial lipopeptides), and scRNA-seq was performed at 1, 2, 4 and 6 h after stimulation. With the capture time information, the original study studied single-cell gene expression dynamics under the two exposures, however, the cells behave \emph{asynchronously} and heterogeneity exists within the cellular populations at each time point (i.e.\ after 1 hour of stimulation, the cells do not reach exactly the same biochemical state). Previous analyses have suggested this data set is more suited to a latent variable (so called \emph{pseudotime}) analysis \citep{campbell_uncovering_2018} where the capture times are treated as unreliable and the latent variable is used instead to capture a continuous measure of how the cells are continually changing in response to each stimuli (Figure \ref{fig:lpspam}). 

\begin{figure*}[!t]
    \centering
    \includegraphics[width=0.8\textwidth]{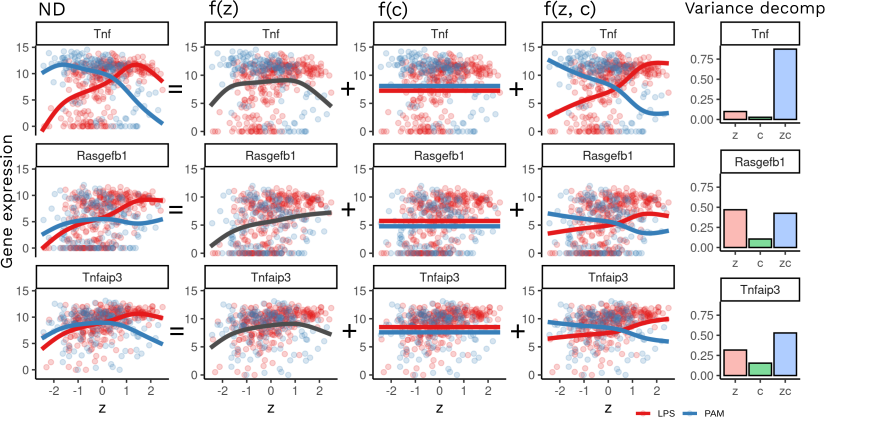}
    \vspace{-3mm}
    \caption{ND on the large-scale scRNA-seq data of dendritic cells, using $c \in \{\text{LPS}, \text{PAM}\}$, lets us decompose gene-level variation (shown for selected 3 genes) into the marginal $z$ effect, the marginal $c$ effect, and the non-linear interaction between the two, together with the fraction of variance attributed to each component. As opposed to earlier linear approaches \citep{campbell_uncovering_2018}, the decomposition provided by ND is non-linear.}
    \label{fig:shalek}
    \vspace{-3mm}
\end{figure*}

We conducted an analysis using ND where we encoded the stimulant to which the cells were exposed as a binary covariate ($c \in \{\text{LPS}, \text{PAM}\}$) and used a single dimension for the latent variable $z$. Each gene was therefore modelled as a combination of main effects due to 1) LPS/PAM exposure, 2) temporal effects (independent of stimulant type), and 3) temporal interaction effects that were modulated by the stimulant. 

We applied ND to 820 cells using the 7,500 most variable genes, following the previously published \emph{linear} analysis \citep{campbell_uncovering_2018}. Our inferred latent dimension was indeed correlated with capture time and therefore is indeed a measure of the continuous progression of these cells under stimulation (Figure \ref{fig:lpspam}). 

Furthermore, we were able to decompose the expression of each gene into components that were dependent on non-linear additive effects of pseudotime $z$, covariate $c$, and interactions ($(z, c)$). In Figure \ref{fig:shalek}, we highlight three genes previously identified to exhibit interaction effects. However, a limitation of the previous analysis was the reliance on linear models, which leaves open the possibility of model misspecification driving false positives in this very noisy single cell expression data. Here we have used a more flexible, non-linear CVAE-based approach. The genes \emph{Tnf}, \emph{Rasgefb1} and \emph{Tnfaip3} all exhibited strong interaction effects and the dependence of the gene expression on $(z, c)$ does follow a near-linear behaviour even under a flexible non-linear model, thus confirming the findings of \citep{campbell_uncovering_2018} under a more flexible class of models, while providing a more accurate feature-level variance decomposition (rightmost panels in Figure~\ref{fig:shalek}).

We next identified the top 50 genes with the strongest 1) additive $z$, 2) additive $c$, and 3) interactions effects as identified by ND, and applied UMAP, a popular dimensionality reduction and visualisation algorithm, to examine these subsets in a two-dimensional representation (Supp Fig 1). As expected, the set of genes with strong additive $z$ effects but smaller additive $c$ and interaction effects, showed considerable intermixing between cells stimulated under the two conditions. These genes exhibit behaviours that are are largely independent of stimulus so we would not expect segregation of the genes by stimulus type. In contrast, the UMAP visualisation of the gene sets with strong additive $c$ or interaction effects shows considerable separation between the LPS and PAM stimulated cells. These are genes whose behaviour is heavily influenced by the type of stimulation used as well as the latent variable. 

Overall, this suggests that ND-CVAE is identifying relevant structure in the single cell data and correctly attributing the appropriate feature behaviours to the relevant subsets of genes. Furthermore, unlike previous analyses, our flexible non-linear models permit greater robustness to model misspecification.

\section{Discussion}

We have proposed a VAE-framework where we embed functional ANOVA decompositions within the decoding structure. This specification allows us to associate variation in the latent space and other covariates to feature-level variability, leading to interpretability that is not present in existing (C)VAE models. Our work brings together ideas from classical statistics and constrained optimisation, while leveraging modern deep learning software to develop comparatively fast, scalable, and interpretable models for dimensionality reduction. 

The ND construction makes the \emph{functional decomposition} (with fixed inputs) identifiable. In principle, the entire model could be made identifiable, when combining ND with monotonicity constraints on the neural networks \citep{pierson_inferring_2019}, however, this would considerably restrict the flexibility of the model.

\subsection*{Acknowledgements}

KM was supported by a UK Engineering and Physical Sciences Research Council Doctoral Studentship. CY is supported by a UK Medical Research Council Research Grant (Ref: MR/P02646X/1) and by The Alan Turing Institute under the EPSRC grant EP/N510129/1.

\bibliography{references}

\clearpage

\onecolumn

\begin{appendices}

\section*{Supplementary Information}

\section{Experimental details}

\subsection{Synthetic data generative mechanism (for Figure 1)}

We used the following data generative mechanism. For $i=1, \ldots, N$ ($N=500$) we generated
\begin{itemize}
    \item $z_i \sim U(-2, 2)$
    \item $c_i \sim U(-2, 2)$
    \item $y_i^{(1)} := \exp(-z_i^2) + 0.3 \text{tanh}(x) + \varepsilon_i$
    \item $y_i^{(2)} := \sin(z_i) + 0.2x_i + 0.2\sin(z_i) \cdot x_i \cdot I(z_i > 0) + \varepsilon_i$
\end{itemize}

\subsection{Synthetic data generative mechanism (for Figure 4)}

\begin{itemize}
    \item $z_i \sim U(-2, 2)$ for $i=1, ..., N$
    \item $c_i \sim U(-2, 2)$ for $i=1, ..., N$
    \item for features $j=1, \ldots 5$ 
    \begin{itemize}
        \item $y^{(j)} := w_j \cdot \cos(z) + \varepsilon$ where $w_j \in \{0.1, 0.2, 0.3, 0.4, 0.5\}$
    \end{itemize}
    \item for features $j=6, \ldots 10$ 
    \begin{itemize}
        \item $y^{(j)} := 0.5 z + w_j c + \varepsilon$ where $w_j \in \{0.05, 0.1, 0.15, 0.2, 0.25\}$
    \end{itemize}
    \item for features $j=11, \ldots 15$
    \begin{itemize}
        \item $y^{(j)} := w_j \tanh(z)c + \varepsilon$ where $w_j \in \{0.01, 0.02, 0.03, 0.04, 0.05\}$
    \end{itemize}
    \item for features $j=16, \ldots 20$ 
    \begin{itemize}
        \item $y^{(j)} := w_j c + 0.01 (0.12-w_j) z \tanh(c) + \varepsilon$ where $w_j \in \{0.02, 0.04, 0.06, 0.08, 0.1\}$
    \end{itemize}
    \item for features $j=21, \ldots 25$ 
    \begin{itemize}
        \item $y_i^{(j)} := 0.1 \tanh(z) + 0.2 \tanh(c) + w_j \sin(z) x + \varepsilon$ where $w_j \in \{0.2 , 0.4, 0.6, 0.8, 1.0\}$
    \end{itemize}
    with noise $\varepsilon_i \sim \mathcal{N}(0, \sigma^2)$ with $\sigma = 0.05$
\end{itemize}

\section{Additional results on single-cell data}

\begin{figure}[!h]
    \centering
    \includegraphics[width=0.75\columnwidth]{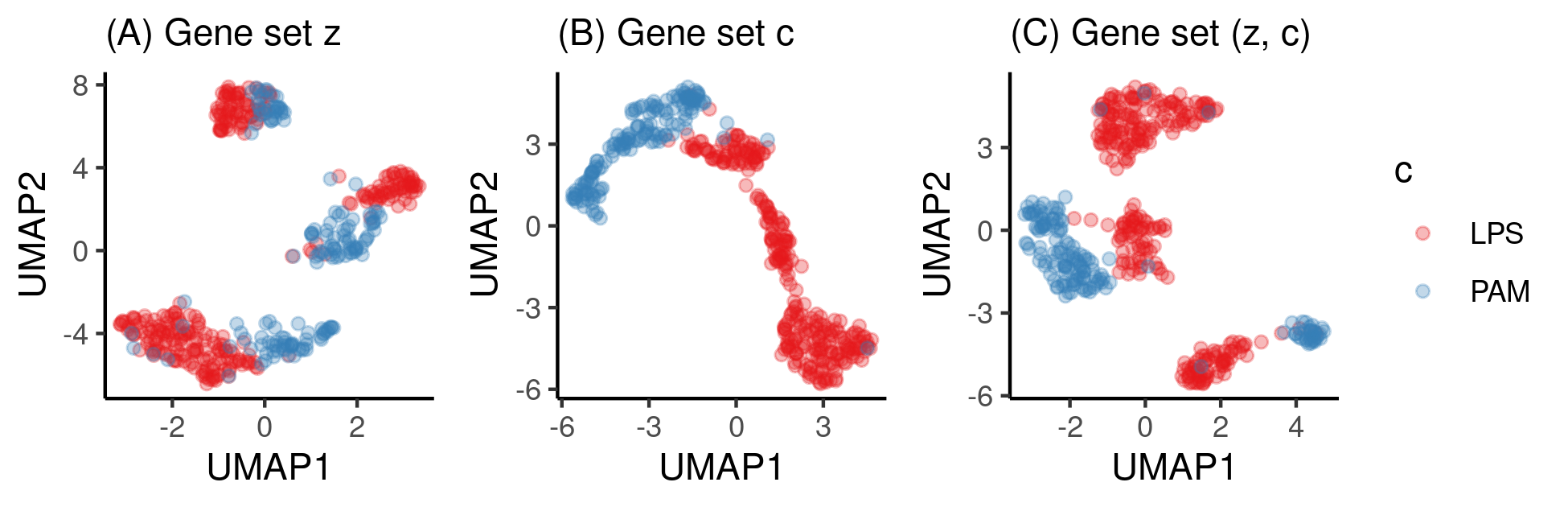}
    \caption{UMAP visualisations of gene sets with large (A) additive $z$, (B) additive $c$ and (C) interaction effects from the mouse dendritic single cell data.}
    \label{fig:shalek_umap}
\end{figure}





\section{Batch correction experiment (two-dimensional latent space)}

Here we devise a synthetic experiment generated from a two-dimensional latent space $(z_1, z_2)$. The data consists of two batches where each feature is either unperturbed, differs by a constant by batch or varies with $z_1$ by batch. 

Our goal was two-fold:
\begin{itemize}
    \item To learn a 2D latent space where $\boldz$ would be adjusted for the confounding batch effect (i.e.\ we want $\boldz$ not to be predictive of the batch label)
    \item To characterise the variance decomposition of every feature in terms of $z_1, z_2$ and $c$. 
\end{itemize}

So we generated data according to the following scheme (where $c_i$ are batch indicators):

\begin{itemize}
    \item $z_{1, i} \sim U(-2, 2)$ for $i=1, ..., N$
    \item $z_{2, i} \sim U(-2, 2)$ for $i=1, ..., N$
    \item $c_i \sim \text{Bernoulli}(0.5)$ for $i=1, ..., N$
    \item for features $j=1, \ldots 5$ 
    \begin{itemize}
        \item $y^{(j)} := 0.3 w_j \tanh(z_1) + 0.2 w_j \exp(-0.5 z_2^2) + 0.3 w_j c + \varepsilon$ where $w_j \in \{1, 2, 3, 4, 5\}$
    \end{itemize}
    \item for features $j=6, \ldots 10$ 
    \begin{itemize}
        \item $y^{(j)} := 0.2 w_j \tanh(z_2) + 0.4 (6-w_j) c + \varepsilon$ where $w_j \in \{1, 2, 3, 4, 5\}$
    \end{itemize}
    \item for features $j=11, \ldots 15$
    \begin{itemize}
        \item $y^{(j)} := w_j z_1 + (0.6-w_j) z_2 + (0.6-w_j) \tanh(z_1) c + \varepsilon$ where $w_j \in \{0.1, 0.2, 0.3, 0.4, 0.5\}$
    \end{itemize}
    \item for features $j=16, \ldots 20$ 
    \begin{itemize}
        \item $y^{(j)} := \tanh(2 z_1) + w_j \tanh(z_2) + \varepsilon$ where $w_j \in \{0.2, 0.4, 0.6, 0.8, 1.0\}$
    \end{itemize}
    \item for features $j=21, \ldots 25$ 
    \begin{itemize}
        \item $y_i^{(j)} := 0.1 c + w_j \tanh(z_1) c + \varepsilon$ where $w_j \in \{0.2, 0.4, 0.6, 0.8, 1.0\}$
    \end{itemize}
    with noise $\varepsilon_i \sim \mathcal{N}(0, \sigma^2)$ with $\sigma = 0.05$ 
\end{itemize}

\begin{figure}[!h]
    \centering
    \includegraphics[width=\textwidth]{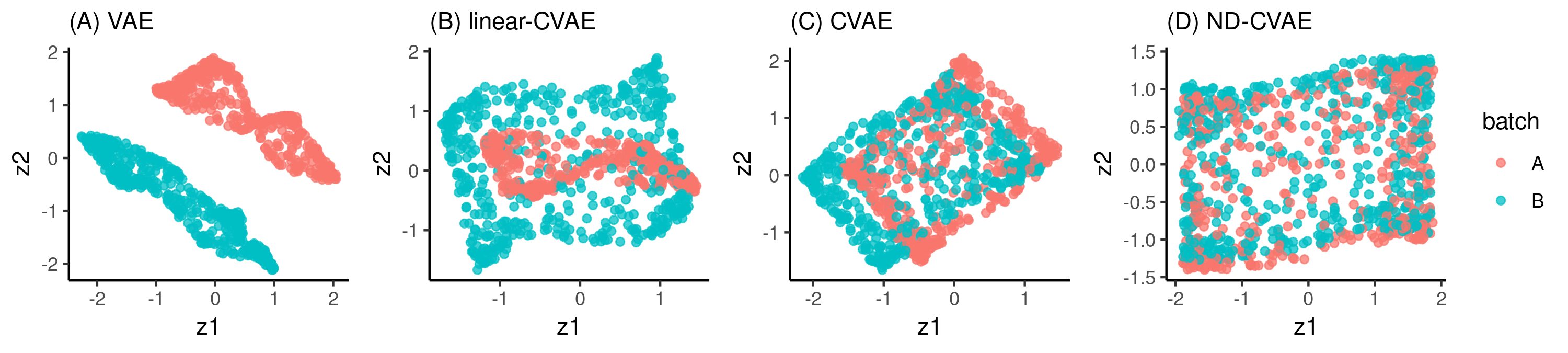}
    \vspace{-1mm}
    \caption{ND-CVAE recovers a batch-adjusted two-dimensional $\boldz$ on a synthetic example whereas other approaches (VAE, CVAE, and linear-CVAE) struggle to appropriately adjust for known $\boldc$.}
    \label{fig:toy_exprs_2D}
\end{figure}

The goal was to identify if any tested VAE variant was capable of achieving \emph{batch correction} (here $c$=[batch]) by identifying a latent space in which the two batches overlapped each other (Figure~\ref{fig:toy_exprs_2D}). 

\begin{figure}[!h]
    \centering
    \includegraphics[width=\textwidth]{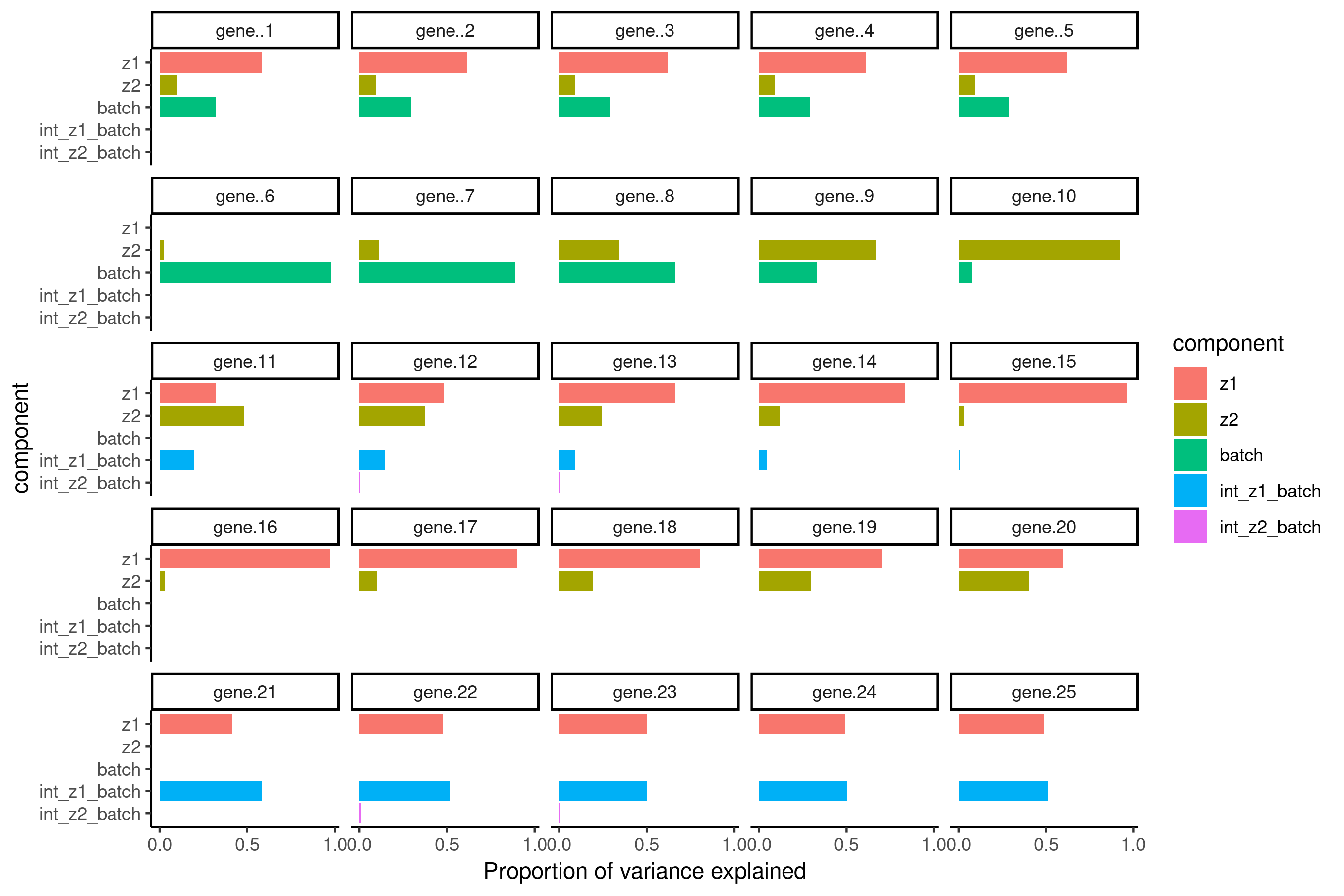}
    \caption{On the synthetic batch-correction example, we characterised the decomposition learned by ND for every gene as a function of $z_1$, $z_2$, batch $c$, and interactions between them. }
    \label{fig:supp_2D_var_decomp}
\end{figure}

With the restricted representation power of Linear-CVAE, only translational shifts in the latent space could be corrected. Surprisingly, the standard CVAE did not entirely remove the batch effect in the latent space either. However, the sparse ND structure within the ND-CVAE has correctly identified a $(z_1, z_2)$ space in which the batches are now intermixed and the nonlinear batch effects removed. Furthermore, ND-CVAE lets us characterise how features vary with latent $z_1, z_2$ and known $c$ (Figure~\ref{fig:supp_2D_var_decomp}).

\newpage
\section{Connection to a GP-based decomposition}

Here we discuss how the Neural Decomposition behaviour relates to a conceptually similar c-GPLVM decomposition \citep{martens_decomposing_2019}. The latter, being a decomposition of Gaussian Processes, is better understood in the sense that for any configuration of kernel hyperparameters, the integral constraints can be fulfilled analytically in closed form via conditioning. Thus, the GP decomposition is \emph{exact} in the sense that all the integral constraints can be fulfilled exactly rather than approximately. 
However, the two model classes have different properties: they make different assumptions and have different scalability properties. We will discuss both of these below. 

\subsection{Enforcing integral constraints}

Thus, in some sense, one could consider a GP-decomposition as a golden standard for this purpose, however note that the set of functions that have positive probability under the GP prior (e.g.\ under the squared exponential kernel these are infinitely differentiable functions) does not necessarily overlap with the set of functions that are parameterised by a neural network. The former is determined by the kernel, whereas the latter is determined by various neural architectural choices.

\begin{figure}[!h]
    \centering
    \includegraphics[width=0.95\textwidth]{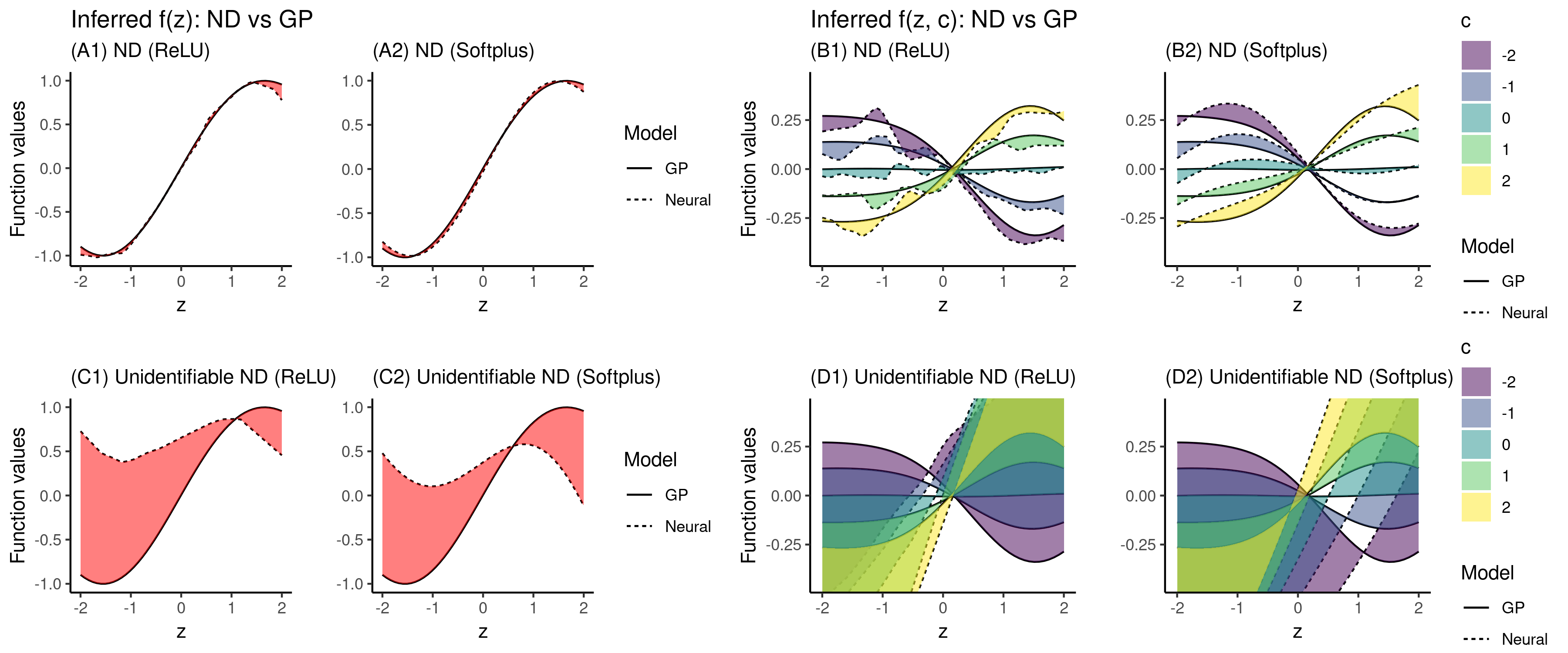}
    \vspace{-2mm}
    \caption{On a synthetic data set ($N=500$) we compare the Neural Decomposition with the GP exact decomposition directly in the function space, visualising both the predicted mean for the inferred $f_z$ (panels A) and $f_{zc}$ (panels B, shown for $c \in \{-2, -1, 0, 1, 2\}$) as well as the $L^1$ distance (shaded area) between them. The functional subspace defined by ND depends on various architectural choices such as nonlinearity: we used ReLU in panels (A1, B1) and Softplus in (A2, B2). The ND without identifiability constraints (bottom row) learns a decomposition which lies far (in $L^1$ distance) from the GP one. }
    \label{fig:L1_comparison_with_GP}
\end{figure}

In Figure~\ref{fig:L1_comparison_with_GP} we have investigated how the behaviour of Neural Decomposition differs from the GP-based functional decomposition on synthetic example that involves inference over $f_z$, $f_c$, and $f_{zc}$. 
We have visualised the inferred GP mean and the inferred ND mappings (using a one hidden layer architecture, with 64 neurons, either with a ReLU or Softplus nonlinearity) for both $f_z$ (on the left) and $f_{zc}$ (on the right), highlighting the $L^1$ distance between the two functions. For the ND with identifiability constraints (top row), this distance is relatively small and there are only minor differences from the GP posterior means, whereas the mappings inferred by the unidentifiable ND without constraints (bottom row) differ significantly from the GP ones. 

\subsection{Computational considerations}

\begin{figure}[!ht]
    \centering
    \includegraphics[width=0.8\textwidth]{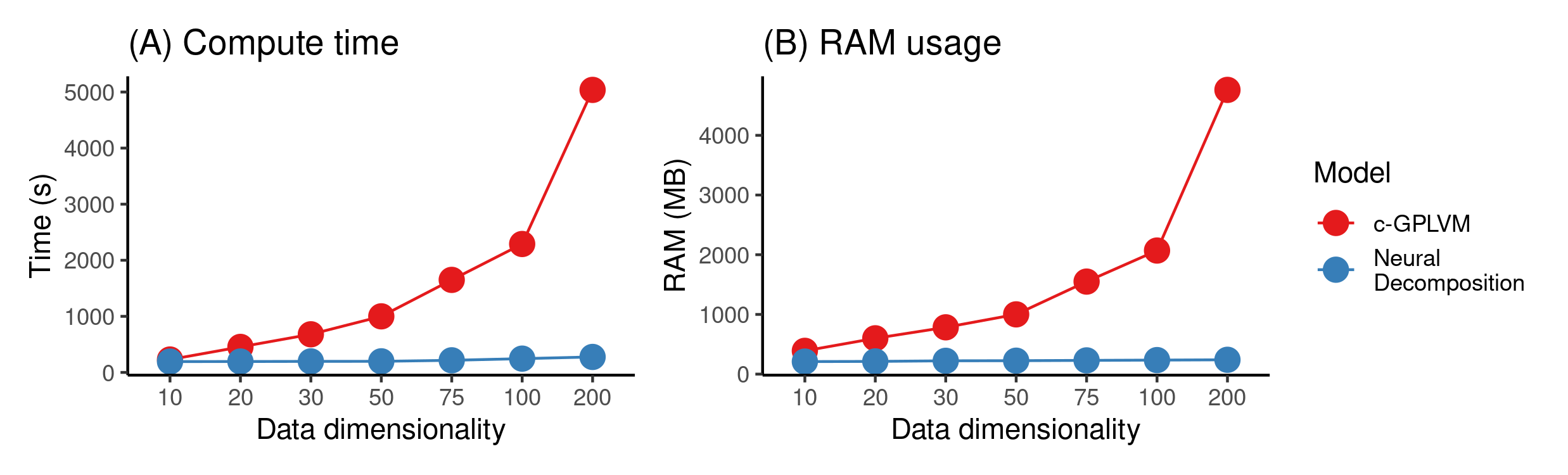}
    \caption{
        Neural decomposition is much more scalable w.r.t.\ data dimensionality than the GP decomposition (inducing-point implementation of the c-GPLVM), both in terms of compute time and memory requirements. For fixed sample size $N=500$, we varied the dimensionality of data $P \in \{10, 20,  \ldots, 100, 200\}$, and compared (A) the compute time, and (B) RAM usage for both methods when varying $P$ (on $x$-axis). 
        Experiments were made on a desktop with 8 Intel i7-6700 CPUs, both implementations in PyTorch.
    }
    \label{fig:RAM_and_time}
\end{figure}

Despite its elegant theoretical underpinnings, the GP-decomposition suffers from scalability issues intrinsic to GP-based models. While the cubic complexity w.r.t.\ sample size $N$ can be addressed via inducing-point methods, the scalability of c-GPLVM w.r.t.\ data dimensionality $P$ can become the limiting factor for high-dimensional data. While the decomposable c-GPLVM scales linearly with $P$ both in terms of compute and memory, even for moderate $P$ it becomes prohibiting to fit c-GPLVM on a laptop because of the memory requirements, as shown in Figure~\ref{fig:RAM_and_time}. 

\end{appendices}

\end{document}

%% file: preamble_maths.tex
\newcommand{\R}{\mathbb{R}}

\newcommand{\boldW}{\mathbf{W}}

\newcommand{\boldb}{\mathbf{b}}

\newcommand{\boldc}{\mathbf{c}}

\newcommand{\bolds}{\mathbf{s}}

\newcommand{\boldy}{\mathbf{y}}
\newcommand{\boldx}{\mathbf{x}}

\newcommand{\boldu}{\mathbf{u}}

\newcommand{\boldz}{\mathbf{z}}

\newcommand{\Var}{\mathrm{Var}}

\newcommand{\calI}{\mathcal{I}}

%% file: main.bbl
\begin{thebibliography}{}

\bibitem[Blei et~al., 2017]{blei_variational_2017}
Blei, D.~M., Kucukelbir, A., and McAuliffe, J.~D. (2017).
\newblock Variational inference: {A} review for statisticians.
\newblock {\em Journal of the American Statistical Association},
  112(518):859--877.

\bibitem[Campbell and Yau, 2018]{campbell_uncovering_2018}
Campbell, K.~R. and Yau, C. (2018).
\newblock Uncovering pseudotemporal trajectories with covariates from single
  cell and bulk expression data.
\newblock {\em Nature communications}, 9(1):2442.

\bibitem[Eraslan et~al., 2019]{eraslan_single-cell_2019}
Eraslan, G., Simon, L.~M., Mircea, M., Mueller, N.~S., and Theis, F.~J. (2019).
\newblock Single-cell {RNA}-seq denoising using a deep count autoencoder.
\newblock {\em Nature communications}, 10(1):390.

\bibitem[Gu, 2013]{gu_smoothing_2013}
Gu, C. (2013).
\newblock {\em Smoothing spline {ANOVA} models}, volume 297.
\newblock Springer Science \& Business Media.

\bibitem[Hestenes, 1969]{hestenes_multiplier_1969}
Hestenes, M.~R. (1969).
\newblock Multiplier and gradient methods.
\newblock {\em Journal of optimization theory and applications}, 4(5):303--320.

\bibitem[Jang et~al., 2017]{jang_categorical_2017}
Jang, E., Gu, S., and Poole, B. (2017).
\newblock Categorical {Reparameterization} with {Gumbel}-{Softmax}.
\newblock {\em International Conference on Learning Representations}.

\bibitem[Kaufman and Sain, 2010]{kaufman_bayesian_2010}
Kaufman, C.~G. and Sain, S.~R. (2010).
\newblock Bayesian functional {ANOVA} modeling using {Gaussian} process prior
  distributions.
\newblock {\em Bayesian Analysis}, 5(1):123--149.

\bibitem[Kim and Mnih, 2018]{kim_disentangling_2018}
Kim, H. and Mnih, A. (2018).
\newblock Disentangling by {Factorising}.
\newblock In {\em International {Conference} on {Machine} {Learning}}, pages
  2649--2658.

\bibitem[Kingma and Ba, 2014]{kingma_adam:_2014}
Kingma, D.~P. and Ba, J. (2014).
\newblock Adam: {A} method for stochastic optimization.
\newblock {\em Proceedings of the 3rd International Conference on Learning
  Representations (ICLR)}.

\bibitem[Kingma and Welling, 2014]{kingma_auto-encoding_2014}
Kingma, D.~P. and Welling, M. (2014).
\newblock Auto-encoding variational bayes.
\newblock {\em Proceedings of the International Conference on Learning
  Representations (ICLR)}.

\bibitem[Lengerich et~al., 2019]{lengerich_purifying_2019}
Lengerich, B., Tan, S., Chang, C.-H., Hooker, G., and Caruana, R. (2019).
\newblock Purifying {Interaction} {Effects} with the {Functional} {ANOVA}: {An}
  {Efficient} {Algorithm} for {Recovering} {Identifiable} {Additive} {Models}.
\newblock {\em arXiv preprint arXiv:1911.04974}.

\bibitem[Liu et~al., 2015]{liu_deep_2015}
Liu, Z., Luo, P., Wang, X., and Tang, X. (2015).
\newblock Deep learning face attributes in the wild.
\newblock In {\em Proceedings of the {IEEE} international conference on
  computer vision}, pages 3730--3738.

\bibitem[Lopez et~al., 2018]{lopez_deep_2018}
Lopez, R., Regier, J., Cole, M.~B., Jordan, M.~I., and Yosef, N. (2018).
\newblock Deep generative modeling for single-cell transcriptomics.
\newblock {\em Nature methods}, 15(12):1053.

\bibitem[Lundberg and Lee, 2017]{lundberg_unified_2017}
Lundberg, S.~M. and Lee, S.-I. (2017).
\newblock A unified approach to interpreting model predictions.
\newblock In {\em Advances in {Neural} {Information} {Processing} {Systems}},
  pages 4765--4774.

\bibitem[Maddison et~al., 2017]{maddison_concrete_2017}
Maddison, C.~J., Mnih, A., and Teh, Y.~W. (2017).
\newblock The {Concrete} {Distribution}: {A} {Continuous} {Relaxation} of
  {Discrete} {Random} {Variables}.
\newblock In {\em International {Conference} on {Learning} {Representations}}.

\bibitem[Märtens et~al., 2019]{martens_decomposing_2019}
Märtens, K., Campbell, K., and Yau, C. (2019).
\newblock Decomposing feature-level variation with {Covariate} {Gaussian}
  {Process} {Latent} {Variable} {Models}.
\newblock In {\em International {Conference} on {Machine} {Learning}}, pages
  4372--4381.

\bibitem[Märtens and Yau, 2020]{martens_basisvae_2020}
Märtens, K. and Yau, C. (2020).
\newblock {BasisVAE}: {Translation}-invariant feature-level clustering with
  {Variational} {Autoencoders}.
\newblock In {\em International {Conference} on {Artificial} {Intelligence} and
  {Statistics} ({AISTATS})}.

\bibitem[Pierson et~al., 2019]{pierson_inferring_2019}
Pierson, E., Koh, P.~W., Hashimoto, T., Koller, D., Leskovec, J., Eriksson, N.,
  and Liang, P. (2019).
\newblock Inferring {Multidimensional} {Rates} of {Aging} from
  {Cross}-{Sectional} {Data}.
\newblock {\em Proceedings of machine learning research}, 89:97.

\bibitem[Platt and Barr, 1988]{platt_constrained_1988}
Platt, J.~C. and Barr, A.~H. (1988).
\newblock Constrained differential optimization.
\newblock In {\em Neural {Information} {Processing} {Systems}}, pages 612--621.

\bibitem[Powell, 1969]{powell_method_1969}
Powell, M.~J. (1969).
\newblock A method for nonlinear constraints in minimization problems.
\newblock {\em Optimization}, pages 283--298.

\bibitem[Ramsay and Silvermann, 1997]{ramsay_functional_1997}
Ramsay, J.~O. and Silvermann, B.~W. (1997).
\newblock {\em Functional {Data} {Analysis}.}
\newblock Springer {Series} in {Statistics}.

\bibitem[Rezende et~al., 2014]{rezende_stochastic_2014}
Rezende, D.~J., Mohamed, S., and Wierstra, D. (2014).
\newblock Stochastic backpropagation and approximate inference in deep
  generative models.
\newblock {\em arXiv preprint arXiv:1401.4082}.

\bibitem[Ribeiro et~al., 2016]{ribeiro_why_2016}
Ribeiro, M.~T., Singh, S., and Guestrin, C. (2016).
\newblock Why should i trust you?: {Explaining} the predictions of any
  classifier.
\newblock In {\em Proceedings of the 22nd {ACM} {SIGKDD} international
  conference on knowledge discovery and data mining}, pages 1135--1144. ACM.

\bibitem[Shalek et~al., 2014]{shalek_single-cell_2014}
Shalek, A.~K., Satija, R., Shuga, J., Trombetta, J.~J., Gennert, D., Lu, D.,
  Chen, P., Gertner, R.~S., Gaublomme, J.~T., and Yosef, N. (2014).
\newblock Single-cell {RNA}-seq reveals dynamic paracrine control of cellular
  variation.
\newblock {\em Nature}, 510(7505):363.

\bibitem[Sobol, 1993]{sobol_sensitivity_1993}
Sobol, I.~M. (1993).
\newblock Sensitivity estimates for nonlinear mathematical models.
\newblock {\em Mathematical modelling and computational experiments},
  1(4):407--414.

\bibitem[Sohn et~al., 2015]{sohn_learning_2015}
Sohn, K., Lee, H., and Yan, X. (2015).
\newblock Learning structured output representation using deep conditional
  generative models.
\newblock In {\em Advances in neural information processing systems}, pages
  3483--3491.

\bibitem[Tan et~al., 2018]{tan_learning_2018}
Tan, S., Caruana, R., Hooker, G., Koch, P., and Gordo, A. (2018).
\newblock Learning global additive explanations for neural nets using model
  distillation.
\newblock {\em arXiv preprint arXiv:1801.08640}.

\bibitem[Zhao et~al., 2018]{zhao_bias_2018}
Zhao, S., Ren, H., Yuan, A., Song, J., Goodman, N., and Ermon, S. (2018).
\newblock Bias and generalization in deep generative models: {An} empirical
  study.
\newblock In {\em Advances in {Neural} {Information} {Processing} {Systems}},
  pages 10792--10801.

\end{thebibliography}
